\documentclass[10pt,twocolumn,letterpaper]{article}

\usepackage{cvpr}
\usepackage{times}
\usepackage{epsfig}
\usepackage{graphicx}
\usepackage{amsmath}
\usepackage{amssymb}
\usepackage{subfigure}

\usepackage[pagebackref=true,breaklinks=true,letterpaper=true,colorlinks,bookmarks=false]{hyperref}
\cvprfinalcopy 

\newcommand{\R}{\mathbb{R}}
\newcommand{\E}{\mathbb{E}}

\newcommand{\x}{\mathbf{x}}
\newcommand{\z}{\mathbf{z}}

\newcounter{savecntr}
\newcounter{restorecntr}

\def\cvprPaperID{1365} 

\ifcvprfinal\fi

\begin{document}

\title{Learning to Manipulate Individual Objects in an Image}

\author{Yanchao Yang\setcounter{savecntr}{\value{footnote}}\thanks{Equal contribution.}\\
UCLA Vision Lab\\
{\tt\small yanchao.yang@cs.ucla.edu}
\and
Yutong Chen\thanks{Work is done during the author's visit at UCLA.}
\setcounter{restorecntr}{\value{footnote}}\setcounter{footnote}{\value{savecntr}}\footnotemark\setcounter{footnote}{\value{restorecntr}}\\
Tsinghua University\\
{\tt\small chen-yt16@mails.tsinghua.edu.cn}
\and
Stefano Soatto\\
UCLA Vision Lab\\
{\tt\small soatto@cs.ucla.edu}
}

\maketitle
\thispagestyle{empty}

\begin{abstract}
We describe a method to train a generative model with latent factors that are (approximately) independent and localized. 
This means that perturbing the latent variables affects only local regions of the synthesized image, corresponding to objects. 
Unlike other unsupervised generative models, ours enables object-centric manipulation, without requiring object-level annotations, or any form of annotation for that matter.
The key to our method is the combination of spatial disentanglement, enforced by a Contextual Information Separation loss, and perceptual cycle-consistency, enforced by a loss that penalizes changes in the image partition in response to perturbations of the latent factors. 
We test our method's ability to allow independent control of spatial and semantic factors of variability on existing datasets, and also introduce two new ones which highlight the limitations of current methods.\footnote{Code available at: https://github.com/ChenYutongTHU/Learning-to-manipulate-individual-objects-in-an-image-Implementation}
\end{abstract}

\vspace{-0.1cm}
\section{Introduction}
\vspace{-0.1cm}

Generative models typically aim to capture the natural statistics while isolating independent factors of variation. This can be beneficial if such factors correspond to variables of interest in tasks to be instantiated {\em post-hoc}, or if the model is to be used for image synthesis where the user wants to independently control the outcome. 
Generative models learned from large image collections, for instance variational auto-encoders (VAEs) or generative adversarial networks (GANs) do isolate independent factors of variation, but those affect the global statistics of the image. 
We are interested in spatially-localized factors of variation, so that manipulation of image statistics can occur at the level of {\em objects}, rather than of the whole image. 
While one could learn conditional generative models, this usually requires annotation of the independent factors. 
We aim to learn spatially and semantically independent latent factors without the need for any annotation. 
We call these {\em object-centric generative factors}.

\begin{figure}[!t]
\begin{center}
\footnotesize{}
  \includegraphics[width=0.38\textwidth]{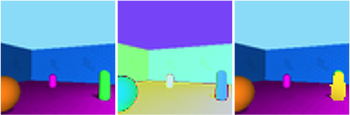}
\end{center}
\vspace{-0.3cm}
\caption{Perturbing the factors learned without knowing objects affects the synthesized scene globally (middle). Object-centric generative factors enable changing the color of the pillar from green (left) to yellow without affecting the other objects (right).}
\vspace{-0.2cm}
\label{fig:global-effect-betaVAE}
\end{figure}

The existing literature on object-centric generative models is restricted to piece-wise constant or smooth images. 
We introduce two variational constraints, one derived from the Contextual Information Separation (CIS) principle \cite{yang2019unsupervised}, but extended to multiple objects, and one derived by enforcing perceptual cycle-consistency, which is that the partition of the image into independently controlled region is stable with respect to perturbations of the latent factors. 
We illustrate the characteristics of our model on existing datasets, and introduce two new datasets of increased complexity.

\vspace{-0.05cm}
\section{Related Work}
\vspace{-0.1cm}

There are two approaches to representation learning, {\em task driven}, where the goal is to learn a function of the data that captures all information relevant to the task and discards everything else (sufficient invariant) \cite{achille2018emergence}, and {\em disentangled}, where the goal is to reconstruct the data while separating the independent factors of variation \cite{bengio2013representation,tran2017disentangled}. 
Technically speaking, the latter is a special case of the former, when the task variable is the data itself, and the independence of latent factor can be framed as the secondary task. 
However, the literature has largely progressed on separate tracks. 
An independent taxonomy can be devised based on the level of supervision.
While disentangled representations usually refer to unsupervised learning, task-driven representations can be unsupervised (if the task is, for instance, prediction, or reconstruction), semi-supervised, or fully supervised. 
Among unsupervised representation learning methods, variational autoencoders (VAE) \cite{kingma2013auto} attempt to extract ``meaningful'' latent factors by forcing a variational bottleneck in the generative model. This can also be seen as a special case of task-driven representation, where the task is the data itself. 
It was shown by \cite{achille2018emergence,press2018emerging} that the latent factors tend to be independent, and can be easily manipulated in a generative setting. 
A different approach uses an adversarial loss \cite{goodfellow2014generative} to map a known distribution (typically a Gaussian) to an approximation of the data distribution, so the input distribution can be considered a set of independent factors. 
To encourage the alignment of the learned representation with the underlying generative factors, several constraints have been proposed to enforce the disentanglement of the latent codes \cite{higgins2017beta,burgess2018understanding,achille2018emergence,press2018emerging,kim2018disentangling,chen2018isolating,bouchacourt2018multi,hsu2017unsupervised}. InfoGAN \cite{chen2016infogan} promotes disentanglement by explicitly maximizing the mutual information between a subset of latent variables and the generated images. Also, domain-specific knowledge could be utilized to learn disentangled representations \cite{tulyakov2018mocogan,tran2017disentangled,karras2019style}. 

In all these methods, disentanglement is sought in latent space, with no grounding on the domain where the data is defined. For the case of images, we would like the independent factors to correspond to compact and simply-connected regions of the image, corresponding to {\em objects}. In all these methods, manipulation of the independent factors typically has global effects on the image, rather than enabling manipulation of one object at a time. We would like to develop models that naturally disentangle the spatial domain, in addition to other latent factors of variation. 

Of course, one could partition the image and independently represent each region, but doing so would require the ability to detect objects in the first place.
Recent methods on structured representation learning have been proposed to enable reasoning about objects in the scene. AIR \cite{eslami2016attend} proposes a structured generative model for the image generation process. 
SQAIR \cite{kosiorek2018sequential} proposes an additional state-space model to enforce temporal consistency such that the decomposition within a sequence is consistent. The capacity of their structured models and the assumption of known/trivial background restrict both AIR and SQAIR. 
DRAW \cite{gregor2015draw} employs attention to generate images with objects in a structured manner. 
In \cite{yao20183d}, an image is decomposed into semantic mask, texture, and geometry, requiring heavy manual supervision.
And \cite{kulkarni2015deep} proposes an auto-encoder that de-renders images into graphics code, which is trained by explicitly specifying the variations.
Similarly, \cite{wu2017neural} employs a graphics engine as the decoder to enforce an interpretable representation, which is, however, not backward-differentiable.
NEM \cite{greff2017neural,van2018relational} construct spatial mixture models to cluster pixels into objects, but only for grayscale images. 
IODINE \cite{greff2019multi} also employs a spatial mixture model to jointly infer the segmentation and object representation. \cite{greff2019multi} does not perform well on textured or cluttered scene, culprit the assumption that pixels can be grouped into objects according to the low dimensional spatial mixture models.

Semantic segmentation has improved considerably since the advent of deep neural networks  \cite{long2015fully,chen2017deeplab,zhang2018context} and so has instance segmentation, where the network has to distinguish between different instances of the same semantic class \cite{romera2016recurrent,hu2017maskrnn,liu2018path,chen2018masklab}.
However, these methods depend on densely annotated ground-truth segmentation masks. 
On the other hand, unsupervised object segmentation has been a long-standing problem, but most of the methods require complicated optimization during inference \cite{yang2015self}. 
Recently, unsupervised learning methods for object segmentation have shown promise: UMODCIS \cite{yang2019unsupervised} proposes contextual information separation for binary moving object detection, with end-to-end training without manual supervision or pseudo masks. 
Later, \cite{arandjelovic2019object} proposes Copy-Pasting GAN to discover binary object masks in images; however, special care has to be taken to prevent trivial solutions.

There is only a handful of work dealing with both segmentation and object-centric representation learning in a unified framework.
Besides the few employing variational inference with a structured model, MONet \cite{burgess2019monet} introduces a recurrent segmentation network within the VAE framework, and trains them jointly to provide segmentation and learned representation.

In the next section, we describe our method and in the following Sect. \ref{sec:experiments} we test the model's ability to capture the statistics of the data while enabling independent control of latent factors corresponding to objects in the scene.

\vspace{-0.05cm}
\section{Method}
\vspace{-0.1cm}
\label{sec:method}

Let $\mathbf{x} \in \R^{H\times W\times 3}$ be a color image, and $\mathbf{z} \in \R^{N}$ be the generative factor of $\mathbf{x}$, which represents different characteristics of the data. 
Our model uses as an inference criterion the Information Bottleneck when the task is the data itself, 
whereby a representation (encoder) $q_{\phi}(\z|\x)$ describes the latent factors (bottleneck) $\z$, and a decoder $p_{\theta}(\x|\z)$ allows sampling images from the latent factors $\z$. 
The encoder and decoder are trained by minimizing the Information Bottleneck Lagrangian (IBL) \cite{achille2018emergence}:
\begin{multline}
\mathcal{L}(\phi, \theta; \x, \beta) = -\E_{\z \sim q_{\phi}(\z|\x)}[\log p_{\theta}(\x|\z)] \\ +\beta\mathbb{KL}(q_{\phi}(\z|\x) \| p(\z))
\label{eq:beta-vae}    
\end{multline}
where $\mathbb{KL}$ is the Kullback–Leibler divergence, and $p(\z)=\mathcal{N}(\mathbf{0},I)$ is usually a zero mean unit variance Gaussian. 
When $\beta = 1$ the IBL reduces to the Evidence Lower Bound (ELBO). For $\beta>1$, \cite{achille2018emergence} shows analytically and  \cite{higgins2017beta} validates empirically, that the latent factors are disentangled. 
However, the factors always affect the generated images globally as shown in Fig. \ref{fig:global-effect-betaVAE}. Our goal here is to infer object-centric generative factors, such that we can perturb the factors associated with a single object, and the perturbation will not affect other objects or entities in the scene. In other words, our goal is to make the generative factors disentangled not only statistically, but also spatially.

\begin{figure}[!t]
\begin{center}
  \includegraphics[width=0.48\textwidth]{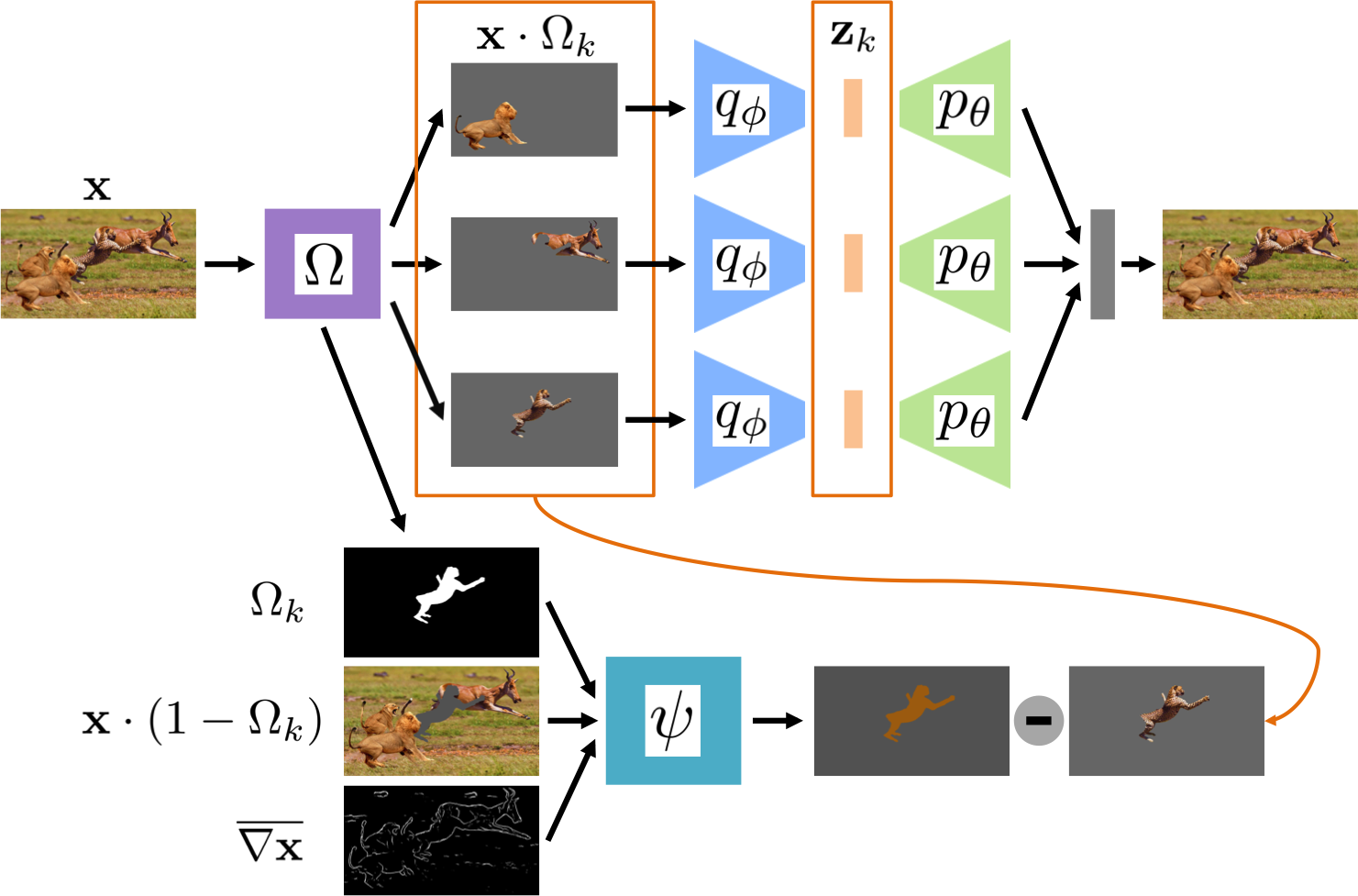}
\end{center}
\vspace{-0.3cm}
\caption{ {\it System Overview.} Our method works by partitioning the image domain into mutually independent regions using the Contextual Information Separation criterion, which entails an inpainting network, and then extracting the generative factors disentangled both spatially and statistically with the identity consistency enforced by the perceptual cycle-consistency constraint. We omit the masks for simplicity.}
\vspace{-0.2cm}
\label{fig:system-overview}
\end{figure}

To this end, we construct a segmentation network to map a color image to $K$ segmentation masks $\Omega_k$'s, with $K$ the maximum number of objects (including background) in the scene:
\begin{equation}
    \Omega: \R^{H\times W\times 3}\rightarrow[0,1]^{H\times W\times K}; \sum_k\Omega(i,j,k)=1, \forall i,j
\end{equation}
Note that $\sum_{i,j}\Omega_k(i,j)=0$ simply means that there is no object in the $k$-th channel. If all the non-zero channels in $\Omega$ represent exactly the segmentation masks of the objects, then we can presumably learn the object-centric generative factors $\z_k$'s as follows:
\begin{equation}
    \mathcal{L}(\phi^\alpha, \phi^s, \theta; \Omega, \x, \beta, \lambda) = \sum_k \mathcal{L}(\phi^\alpha, \phi^s, \theta; \Omega_k, \x, \beta, \lambda)
\label{eq:object-centric-beta-VAE-sum}
\end{equation}
with
\begin{multline}
    \mathcal{L}(\phi^\alpha, \phi^s, \theta; \Omega_k, \x, \beta, \lambda) = \\
    -\E_{ \z_k \sim q_{\phi^\alpha}\cdot q_{\phi^s} } \log p_{\theta}( [\x\cdot\Omega_k,\Omega_k] \mid \z_k ) \\ 
    + \beta\mathbb{KL}( q_{\phi^\alpha}(\z_k^{\alpha} \mid \x\cdot\Omega_k) \| p(\z_k^\alpha)) \\
    + \lambda\mathbb{KL}( q_{\phi^s}(\z_k^s \mid \Omega_k) \| p(\z_k^s))
\label{eq:object-centric-beta-VAE}
\end{multline}
where $q_{\phi^\alpha}$ and $q_{\phi^s}$ are the encoders for appearance and shape related factors of objects in $\x$ respectively. 
Then the joint decoder $\theta$ reconstructs the appearance and mask for each object using $\z_k = \{ \z_k^\alpha, \z_k^s \}$, which is the union of the appearance and shape related factors.
Note that, Eq. \eqref{eq:object-centric-beta-VAE-sum} is a summation of the Information Bottleneck Lagrangians Eq. \eqref{eq:object-centric-beta-VAE} defined on individual segments or ``objects''.
A similar loss is also used in \cite{burgess2019monet} to learn object related representations in an unsupervised manner. However, a question arises from Eq. \eqref{eq:object-centric-beta-VAE-sum}: {\it Why would minimizing the above loss yield a segmentation network $\Omega$ that partitions the image domain into objects?} Given a small enough encoding capacity, it may be true that $\Omega$ will be biased to partition the image $\x$ into pieces that are easier to encode and decode than the full image, while minimizing the first term in Eq. \eqref{eq:object-centric-beta-VAE}, which represents the reconstruction error. 
Then $\Omega$ may succeed when objects happen to be constant color blobs, as they appear in some datasets, \eg, Multi-dSprites and Objects Room~\cite{multiobjectdatasets19}, used for experiments in \cite{burgess2019monet,greff2019multi}. However, what if we want to apply our method on textured objects or cluttered scenes, which are simply not color blobs?

\begin{figure}[!t]
\begin{center}
  \includegraphics[width=0.4\textwidth]{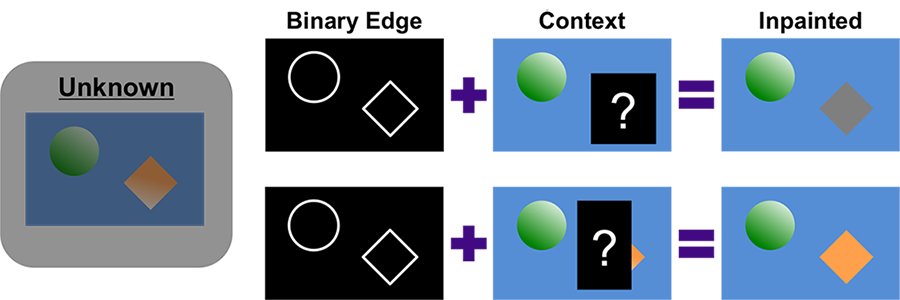}
\end{center}
\vspace{-0.3cm}
\caption{ {\it The average inpainting error using context conditioned on binarized edge map is a good measure of the contextual information computed with the edge conditionals}: Given the binary edge map of an unknown image, the average inpainting error of the masked out region (question mark) will be larger when the context contains less mutual information (the first row), smaller with more mutual information in the context (second row).}
\vspace{-0.3cm}
\label{fig:edge-based-CIS}
\end{figure}

{\bf Spatial Disentanglement.} To endow our method with the ability to learn object-centric generative factors in realistic scenarios, we adapt the Contextual Information Separation (CIS) criterion of \cite{yang2019unsupervised}, which obviates the shortcomings of Eq. \eqref{eq:object-centric-beta-VAE}. 
Instead of a binary segmentation, we extend CIS to multiple objects with the number of objects unknown, and also combine it with the representation learning loss in Eq. \eqref{eq:object-centric-beta-VAE-sum}, which in turn imposes additional regularization through the representational bottleneck.
This way, statistical and spatial disentangling of the generative factors occur simultaneously during learning.

The basic idea of CIS is that, when the context contains no information about a sub-region of an image, the (conditional) reconstruction or ``inpainting'' error will be maximized, as shown in Fig. \ref{fig:edge-based-CIS}. 
In order to measure the mutual information, a joint distribution between a region and its context has to be specified. Here, we choose to use the conditional distribution $p(\x \mid \overline{\nabla\x})$ of an image $\x$ on the binarized edge $\overline{\nabla\x}$. Note that, one could also use the marginal distribution of images $p(\x)$, which may result in degraded performance in general since the mutual information between pixels computed using $p(\x)$ depends on the spatial proximity instead of the structure of the scene. By instantiating a conditional inpainting network $\psi$, our CIS-based spatial disentanglement loss becomes:
\begin{equation}
    \mathcal{L}_{SD}(\psi; \Omega, \x) = \sum_k \dfrac{
    \langle  \Omega_k, \|\psi(\Omega_k, \x\cdot(1-\Omega_k); \overline{\nabla\x})-\x\| \rangle }{\langle \Omega_k, \|\x\| \rangle + \epsilon }
\label{eq:CIS-disentangle-loss}
\end{equation}
where $\langle \cdot, \cdot \rangle$ represents the dot product, and $\|\cdot\|$ is the element-wise $L_1$ norm, which keeps the dimension of the input, $\epsilon$ is a small positive constant that prevents division by zero. 
In \cite{yang2019unsupervised}, it is shown that, under the assumption of Gaussian conditionals, the mutual information can be approximated with the inpainting error.
Note the conditional inpainting network $\psi$ takes in the condition $\overline{\nabla\x}$ and the context $\x\cdot(1-\Omega_k)$ of the image being masked out by $\Omega_k$, and outputs the inpainted image. 
If $\Omega_k$'s perfectly separate the context from each object, this spatial disentanglement loss will be maximized, thus minimizing the mutual information between the inside and outside of $\Omega_k$'s.

{\bf Perceptual Cycle-Consistency.} Given that the decoder $p_{\theta}$ generates images from the object-centric generative factors $\z_k$'s, we can perturb the factors of an appointed object $\hat{\z_k} \sim \mathcal{N}(\z_k, I)$ (Eq. \eqref{eq:PCL-perturb}), 
and synthesize the perturbed image $\hat{\x}$ with $\{ \z_k\}_{\{1,..K\}\backslash k} \cup \hat{\z_k}$ (Eq. \eqref{eq:PCL-syn}), 
and then extract the object-centric generative factors of the perturbed image (Eq. \eqref{eq:PCL-seg},\eqref{eq:PCL-disentangle}). 
If not only the factors are well disentangled statistically and spatially, but also the identities of the disentangled factors are robust to local perturbations, we would expect that the factors extracted from the perturbed image will be unchanged in $\z_k$'s of the other objects, and, also synchronize well with $\hat{\z_k}$, which suggests the following perceptual cycle-consistency loss to further promote disentanglement and identity consistency:
\begin{align}
    k &\sim \mathrm{Uniform(1,K)} \\
    \hat{\z_k} &\sim \mathcal{N}(\z_k, I)
    \label{eq:PCL-perturb}\\
    \hat{\x}   &\leftarrow p_{\theta} ( [\x,\Omega]|\{ \z_k\}_{\{1,..K\}\backslash k} \cup \hat{\z_k} )
    \label{eq:PCL-syn}\\
    \hat{\Omega} &= \Omega(\hat{\x})\label{eq:PCL-seg}\\
    \{ \bar{\z_k} \} &\xleftarrow[]{q_{\phi^\alpha}, q_{\phi^s}} \hat{\x}, \hat{\Omega}
    \label{eq:PCL-disentangle}\\
    \mathcal{L}_{PC}(\phi^{\alpha}, \phi^s, \theta, \Omega, \x ) &=\sum_{l\neq k}\| \z_l - \bar{\z_l} \| + \| \hat{\z_k} - \bar{\z_k} \|
    \label{eq:perceptual-consistency-loss}
\end{align}
Note this characteristic is also desired when we need to track the status of different objects for temporal analysis. 
By combining Eq. \eqref{eq:object-centric-beta-VAE-sum}, \eqref{eq:CIS-disentangle-loss} and \eqref{eq:perceptual-consistency-loss} we have the final training loss for our model:
\begin{multline}
    \arg\max_{\psi}\min_{\phi^\alpha, \phi^s, \theta, \Omega} \mathcal{L}(\phi^\alpha, \phi^s, \theta; \Omega, \x, \beta, \lambda) \\ 
    - \gamma\mathcal{L}_{SD}(\psi; \Omega, \x) + \eta\mathcal{L}_{PC}(\phi^{\alpha}, \phi^s, \theta, \Omega, \x )
\label{eq:final-training-loss}
\end{multline}
Note that, the segmentation network $\Omega$ appears now in three terms, which encourage $\Omega$ to partition the image (first term) while minimizing contextual information (second term) to prevent over-segmentation; \ie, pixels belonging to the same object should be grouped together. Moreover, it has to be robust to perturbations introduced by the third term, which not only imposes perceptual consistency, but can also prevent identity switching; \ie, the object assigned to the $k$-th mask $\Omega_k$ (identified as $k$) should be assigned to $\Omega_k$ again after the perturbations, especially the spatial ones. 
This is particularly useful for applications involving video, since temporal consistency will be automatically achieved after training, and we will show its effectiveness in the experimental section.


\begin{figure}[!t]
\begin{center}
  \includegraphics[width=0.3\textwidth,trim={0 0.1cm 0 2.4cm},clip]{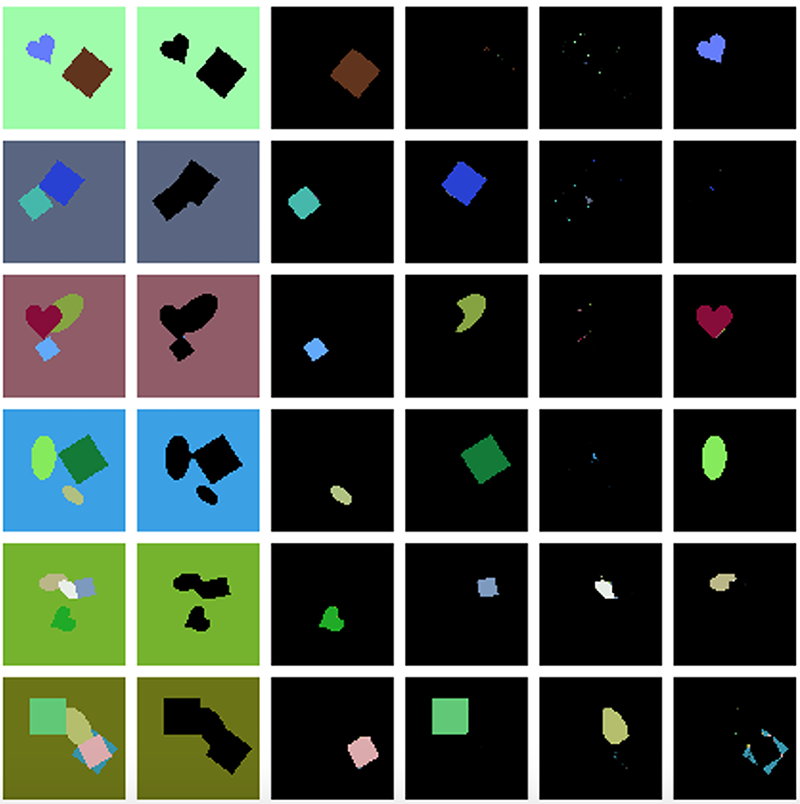}
\end{center}
\vspace{-0.3cm}
\caption{ {\it Spatial Disentanglement on Multi-dSprites}: Our method can segment images containing various numbers of constantly colored objects with heavy occlusions (last two rows).}
\vspace{-0.2cm}
\label{fig:multi_dsprite_seg}
\end{figure}

\begin{figure}[!t]
\begin{center}
  \includegraphics[width=0.35\textwidth,trim={0.05cm 2.0cm 0 0},clip]{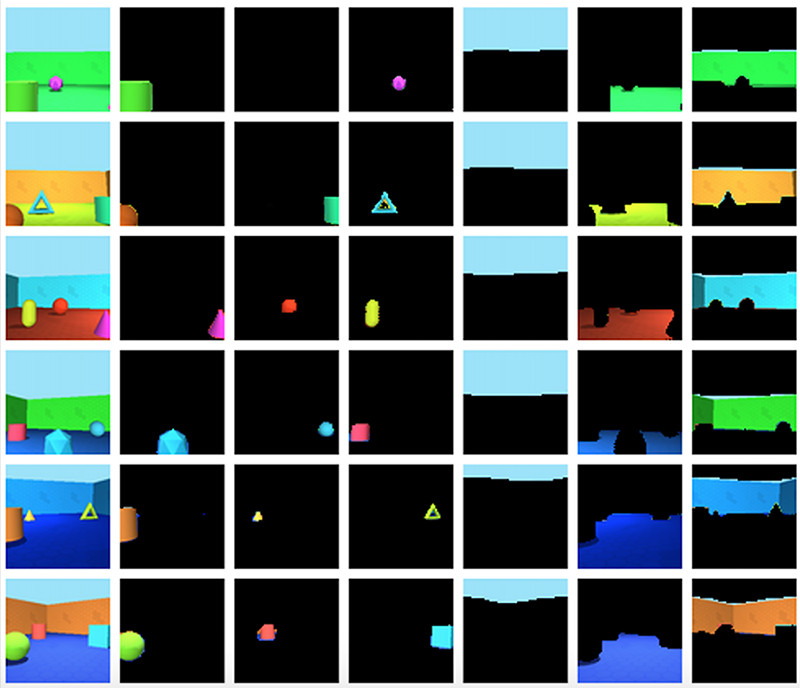}
\end{center}
\vspace{-0.3cm}
\caption{ {\it Spatial Disentanglement on Objects Room}: Our method works on 3D scenes with smoothly colored objects, complex shapes, and different lighting conditions.}
\vspace{-0.2cm}
\label{fig:objects_room_seg}
\end{figure}

\begin{figure}[!ht]
\begin{center}
  \includegraphics[width=0.4\textwidth,trim={0.3cm 0 0 0},clip]{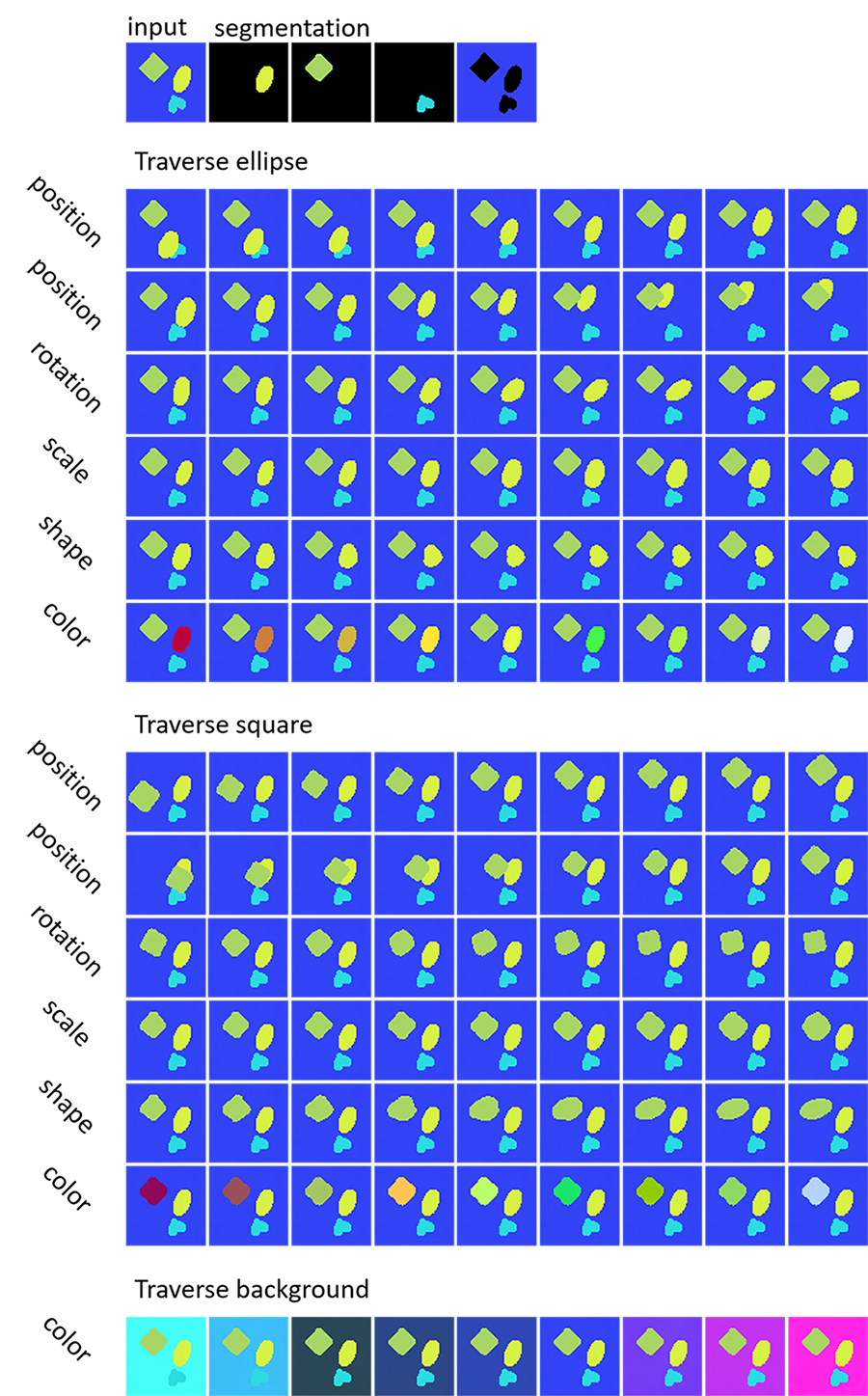}
\end{center}
\vspace{-0.3cm}
\caption{  {\it Traversing the object-centric generative factors on Multi-dSprites}: the input image and the corresponding spatial disentanglement (first row). The rest: traversing the statistically disentangled generative factors of a specific object. Note that the perturbations only affect the object been targeted, and the color of the background can also be modified.}
\vspace{-0.3cm}
\label{fig:multi_dsprites_dis}
\end{figure}

\vspace{-0.05cm}
\section{Experiments}
\vspace{-0.1cm}
\label{sec:experiments}

We first describe the datasets used for evaluation and then elaborate on the implementation details and the training procedure, after which qualitative and quantitative comparisons are provided.

\subsection{Datasets}
\vspace{-0.1cm}

{\bf Multi-dSprites}: dSprites~\cite{matthey2017dsprites} consists of binary images of a single object that varies in shape (square, ellipse, heart), scale, orientation, and position. In Multi-dSprites~\cite{multiobjectdatasets19}, 1-4 shapes are randomly selected from dSprites, randomly colorized and composed on a randomly colored background with occlusions and independent variations in position, scale, and rotation; see sample images in Fig. \ref{fig:multi_dsprite_seg}.

\begin{figure}[!ht]
\begin{center}
  \includegraphics[width=0.35\textwidth,trim={0.4cm 0.1cm 0 0.2cm},clip]{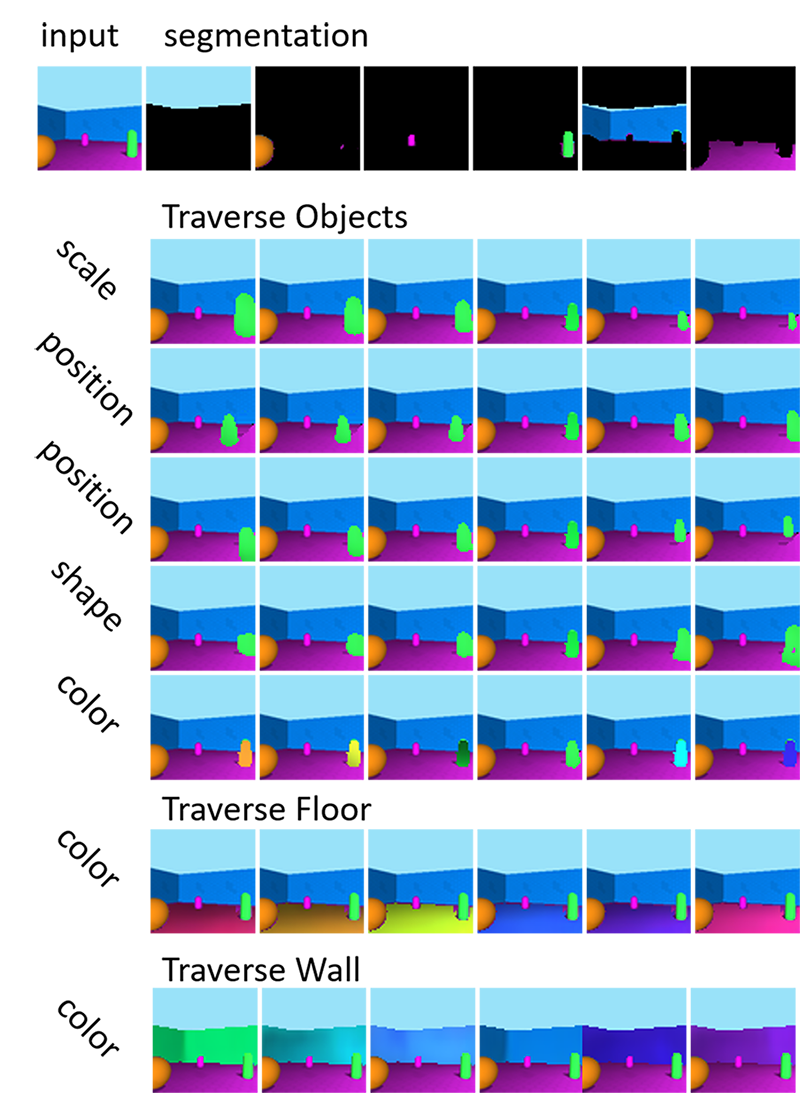}
\end{center}
\vspace{-0.3cm}
\caption{ {\it Traversing the object-centric generative factors on Objects Room}: We can change the scale, position, and color of the green pillar continuously without affecting the others. Also, its shape can deform from a circle to a triangle.}
\vspace{-0.2cm}
\label{fig:objects_room_disentangle}
\end{figure}

\begin{figure}[!h]
\begin{center}
  \includegraphics[width=0.35\textwidth,trim={0 2.6cm 0 0},clip]{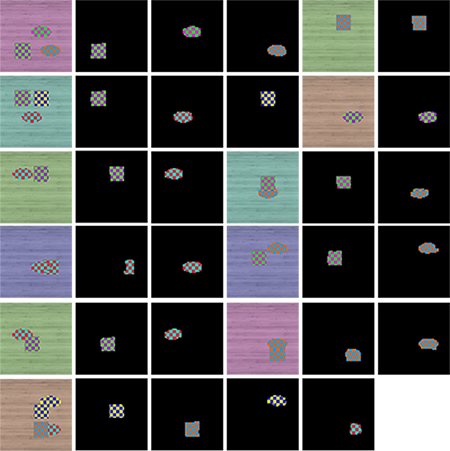}
  \includegraphics[width=0.35\textwidth,trim={0 0 0 6.4cm},clip]{figures/Tex-dSprites-seg-1.png}
\end{center}
\vspace{-0.3cm}
\caption{{\it Spatial Disentanglement on Multi-Texture}: with explicit spatial disentanglement via contextual information separation, our method can segment objects with complex textures.}
\vspace{-0.2cm}
\label{fig:Tex_dSprites_seg}
\end{figure}

{\bf Objects Room} \cite{multiobjectdatasets19} contains rendered images of 3D scenes consisting of 1 to 3 randomly chosen 3D objects that vary in shape, color, size, and pose independently. The wall and the floor of the 3D scene are also colorized randomly. Once projected, objects can exhibit significant appearance variability, depending on lighting and viewpoint. Examining the images from Objects Room in Fig. \ref{fig:objects_room_seg}, we can still see that the images are far from realistic, even though the objects are not uniformly colorized.

{\bf Multi-Texture}: To test whether our proposed method works on complex appearance, for example, textured objects, we create the Multi-Texture dataset. To generate this dataset, 1 to 4 shapes are randomly selected, and independently textured using randomly colorized chessboard patterns. Then, these textured objects are randomly placed on a randomly colorized wooden texture background (Fig. \ref{fig:Tex_dSprites_seg}). 

\begin{figure*}[!ht]
    \centering
    \includegraphics[width=0.75\linewidth,trim={0 2.6cm 0 0},clip]{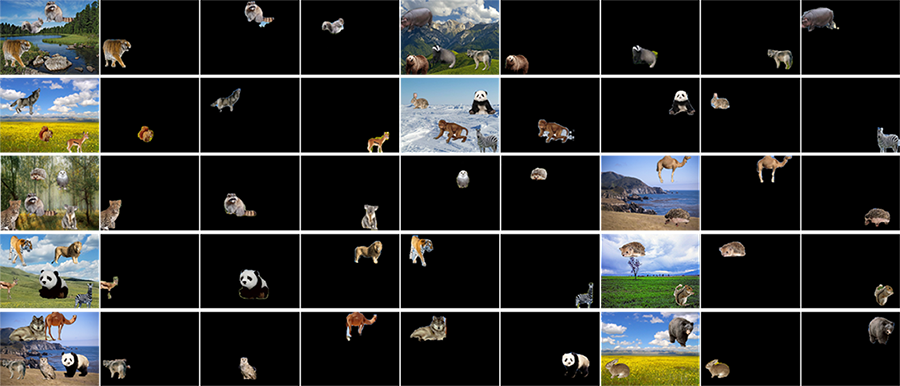}
    \vspace{-0.0cm}
    \caption{ {\it Spatial Disentanglement on Flying Animals.} Our method can spatially disentangle images with natural statistics, where the objects and background are highly non-homogeneous, and also the shape of the objects exhibits more variations than squares or ellipses.}
    \vspace{-0.3cm}
    \label{fig:flying_animal_seg}
\end{figure*}

{\bf Flying Animals}: Although the Multi-Texture is more complex in the object appearance compared to the Objects Room dataset, the homogeneously textured objects still look unnatural in the image statistics, far from the images that would be seen in the real world. For this reason, we come up with the Flying Animals dataset. We collect two sets of natural images. One contains background images from 10 different landscapes, e.g., mountain, desert, and forest, each with 10 different instances; the other set contains clean foreground images of 24 different kinds of animals, each with 10 different instances. We select 1 to 5 objects, randomly scale and position them on a random background image with occlusions. Moreover, we perturb the intensity of each component to simulate different lighting conditions. For sample images please refer to Fig. \ref{fig:flying_animal_seg}.

\begin{figure}[!t]
\begin{center}
  \includegraphics[width=0.35\textwidth]{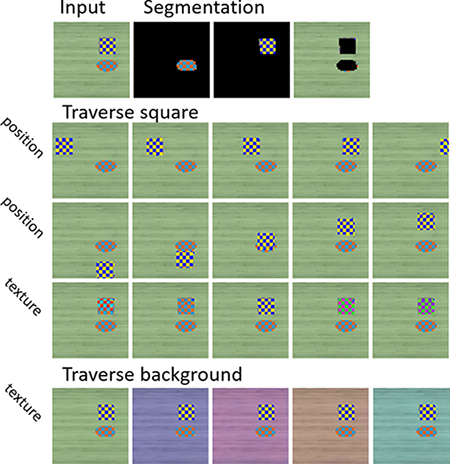}
\end{center}
\vspace{-0.3cm}
\caption{ {\it Traversing the object-centric generative factors on Multi-Texture}: Although the objects are not constantly colored, disentangled object-centric generative factors can still be learned with explicit modeling of the spatial disentanglement.}
\vspace{-0.2cm}
\label{fig:Tex_dSprites_dis}
\end{figure}

\vspace{-0.05cm}
\subsection{Training Details}
\vspace{-0.1cm}

{\bf Segmentation Network $\Omega$}: Similar to the DeepLabV2 architecture \cite{chen2017deeplab}, we use ResNet50 \cite{He2016IdentityMI} as the backbone for our segmentation network, which is followed by four dilated convolution layers in parallel, whose responses are aggregated to generate the K segmentation masks. The total number of trainable parameters is 24M. 

{\bf Inpainting Network $\psi$}: We adapt the inpainting network from \cite{yang2019unsupervised}. It consists of two symmetric encoders that encode binarized edge map and masked image (context), respectively; and a joint decoder with skip-connections from the two encoders. The total number of parameters in the inpainting network is 13M.

{\bf Encoder and Decoder $\phi^\alpha, \phi^s, \theta$}: We adapt the VAE structure proposed in \cite{kim2018disentangling}. 
Instead of a single encoder for images, we instantiate two symmetric encoders $\phi^s$ and $\phi^\alpha$, where $\phi^s$ encodes the one-channel object mask which is the output of the segmentation network, and $\phi^\alpha$ encodes the masked object to get the appearance-related generative factors.
The decoder takes in the object-centric generative factors and generates the objects' appearance and masks, which are then concatenated and fused through four convolutional layers with relu and sigmoid activations to synthesize images over the whole image domain.
The total number of parameters in our encoder-decoder is 1.7M.

{\bf Training}: Adam is used \cite{kingma2014adam} for all modules with initial learning rate 1e-4, epsilon 1e-8, and beta (0.9,0.999). 
As in \cite{yang2019unsupervised}, we find that a pretrained inpainting network will stabilize the training. We randomly crop the input images using rectangular masks with varying height, width, and position, and train the inpainting network to minimize the inpainting error ($L_1$) within the masked region. The training stops after 50K steps.
Then, we update the segmentating network and the inpainting network adversarially to speed up the spatial disentanglement before the joint training of all modules that is performed in an adversarial manner as shown in Eq. \eqref{eq:final-training-loss} and stops after 4M iterations. 
The capacity constraint is adjusted during training following the scheme proposed in \cite{burgess2018understanding}.


\vspace{-0.05cm}
\subsection{Results}
\vspace{-0.1cm}

The closest method to ours is MONet~\cite{burgess2019monet}, which is, to the best of our knowledge, the only one to learn segmentation and representation in a unified framework for non-constantly colored objects. Since we do not have access to the native implementation, we re-implemented MONet by training the same K-way segmentation network in our framework but using the same loss as in \cite{burgess2019monet}. This also eliminates the structural bias that could prevent a fair comparison. Note that we set $K=6$, which is larger than the maximum number of objects that could appear in the training of MONet. In the following, we show the segmentation and learned object-centric generative factors on each dataset. We will also show quantitative evaluations of the jointly learned object segmentation masks.

{\bf Multi-dSprites}: As shown in Fig. \ref{fig:multi_dsprite_seg}, our approach manages to separate different objects such that the VAE can learn representations for every single object and background. Note that the unsupervised segmentation works well with an unknown number of objects and heavy occlusions. Given that spatial disentanglement is achieved through segmentation, we forward each masked object and background into the encoder-decoder and obtain the object-centric factors at the bottleneck. We observe that some dimensions diverge from the prior Gaussian distribution during the training process. As explained in \cite{burgess2018understanding}, these dimensions exhibit semantic meanings aligned with the independent generative factors of dSprites. Fig. \ref{fig:multi_dsprites_dis} displays the segmentation and object-centric disentanglement for an image with three objects. Objects can be manipulated one at a time by perturbing one's latent factors while keeping other objects' representation unchanged. For each object, we can control its independent factors, including positions along two orthogonal axes, rotation, scale, shape, and color, by traversing one dimension in the latent space one at a time.

\begin{figure}[!t]
    \centering
    \includegraphics[width=0.9\linewidth]{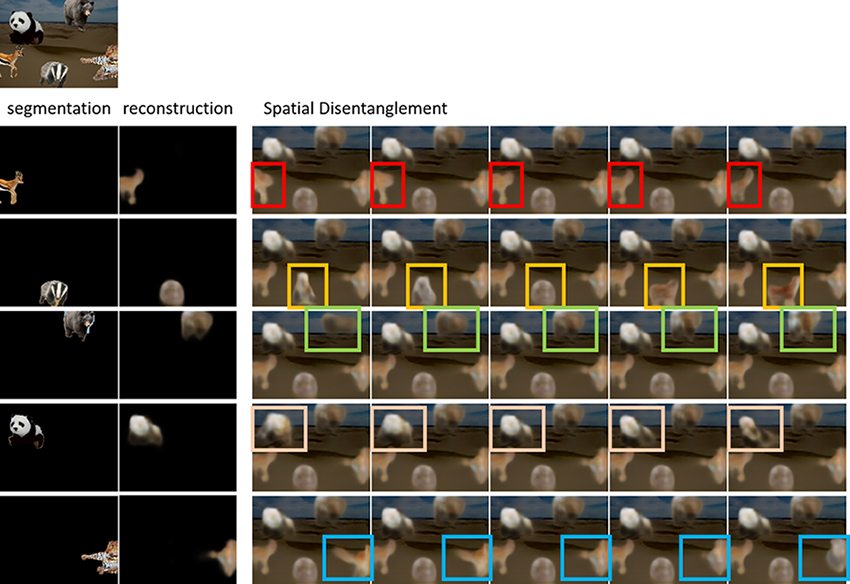}
    \caption{ {\it Traversing the object-centric generative factors on Flying Animals}: The spatial disentanglement of the input image (top-left) is shown in the first column and the second column displays the reconstructed objects by the decoder. The other columns show the traversal on each object. Still, we can change the shape or appearance of each spatially disentangled object individually. For example, in the second row, when perturbing the badger's representation, the appearance of the animal in the yellow box interpolates from owl-like to fox-like while the other four animals and the background remain unchanged.}
    \vspace{-0.3cm}
    \label{fig:flying_animal_disentangle}
\end{figure}

{\bf Objects Room}: Our approach also performs well on the Objects Room dataset, even with shading effects on different objects under various lighting conditions, as shown in Fig. \ref{fig:objects_room_seg}. Object-centric statistical disentanglement is presented in Fig. \ref{fig:objects_room_disentangle}. Similarly, we can edit the scene by changing the position, shape, and color of different objects individually, which shows the applicability on 3D scene editing.

{\bf Multi-Texture}: We experiment on the Multi-Texture dataset to demonstrate that our proposed method enables spatial disentanglement on textured images. As shown in Fig. \ref{fig:Tex_dSprites_seg}, our approach can accurately segment out squares and ellipses with chessboard texture, confirming that the  Contextual Information Separation constraint prevents the network from naively splitting the chessboard into two different colors which correlate with each other. Fig. \ref{fig:Tex_dSprites_dis} shows the disentanglement and the object-centric manipulation results. Even with complex textured objects, our method still enables learning the factors that allow us to change the object consistently, including the background.

{\bf Flying Animals}: To verify that our method is not restricted to synthesized images but can also deal with natural ones, we further test our method on the Flying Animals dataset with real landscapes and animals. As shown in Fig. \ref{fig:flying_animal_seg}, even with complex appearance and shape, our approach can again segment out animals from the natural landscapes, which is far more challenging than segmenting uniformly colored objects as in Multi-dSprites and Objects Room. Similarly, our method can learn disentangled representations for every single animal in the scene and then edit the animals one at a time, shown in Fig. \ref{fig:flying_animal_disentangle}. However, due to the complexity of the appearance and shape, and the trade-off between reconstruction quality and bottleneck capacity for disentanglement, the $\beta$-VAE framework is not powerful enough to conduct statistical disentanglement for each animal while maintaining the details of the object. We will discuss this further in the next section.

\begin{table}
\begin{center}
\footnotesize{}
    \hskip-.32cm
    \begin{tabular}{|c|c|c|c|c|}
    \hline
    Dataset & M-dSprites & Obj-Room & M-Texture & F-Animals \\
    \hline
    MONet & $0.84\pm6.4\delta$ & $0.80\pm8.3\delta$ &  $0.37\pm0.3\delta$ & $0.18\pm2.8\delta$ \\
    \hline
    Ours  & $0.92\pm6.6\delta$ & $0.85\pm5.6\delta$ & $0.88\pm2.6\delta$ & $0.81\pm5.5\delta$ \\
    \hline
    \end{tabular}
    \vspace{-0.0cm}
\caption{ {\it Quantitative evaluation of the segmentation quality between MONet \cite{burgess2019monet} and our method.} Performance measured in the mean intersection-over-union score, reported with mean and variance, where $\delta=10^{-3}$. MONet performs well on Multi-dSprites and Objects Room (constantly or smoothly colored), but its performance drops significantly on Multi-Texture and Flying Animals (textured or complex natural appearance). Our method performs robustly well across different datasets.}
\vspace{-0.4cm}
\label{tb:quantitative}
\end{center}
\end{table}


\begin{table}
\begin{center}
\footnotesize{}
    \begin{tabular}{|c|c|c|}
    \hline
    Perceptual Cycle-Consistency & No & Yes \\
    \hline
    rate of identity switching & 21/255 &  0/255 \\
    \hline
    \end{tabular}
    \vspace{0.0cm}
\caption{ {\it Quantitative evaluation on identity switching}: Note the identity switching decreases to 0 out of 255 by imposing the perceptual cycle-consistency constraint.}
\vspace{-0.6cm}
\label{tb:quantitative_cyc}
\end{center}
\end{table}

\begin{figure}[!t]
    \centering
    \includegraphics[width=0.9\linewidth]{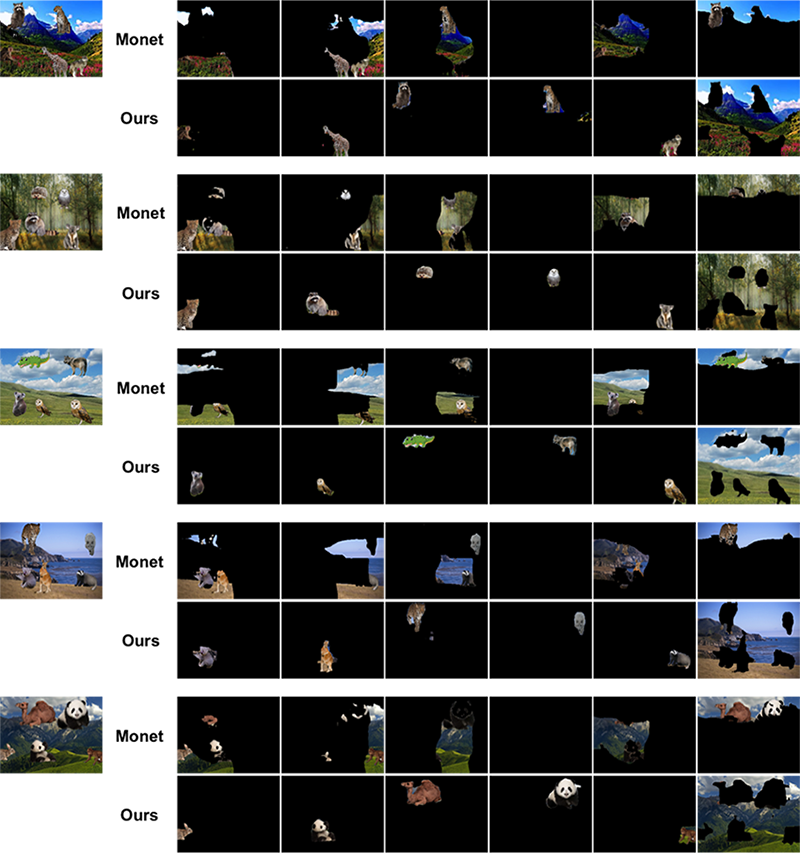}
    \caption{{\it Comparison of the segmentation} between our approach and MONet, which fails to capture objects with natural statistics due to the lack of explicit spatial disentanglement.}
    \vspace{-0.2cm}
    \label{fig:cmp_Monet}
\end{figure}

{\bf Quantitative Evaluation}:
We compare our method with MONet~\cite{burgess2019monet} re-implemented on the four datasets mentioned above. We report the mean-intersection-over-union score (mean$\pm$variance). As shown in Table \ref{tb:quantitative}, our approach achieves better scores than MONet on all four datasets. Particularly, in Multi-Texture and Flying Animals, where objects have complex texture, without explicit information separation (CIS) constraint, MONet tends to segment the images based mainly on color information naively. At the same time, CIS enables our model to ``see'' objects as entities with different parts correlated with each other. To illustrate this, we present the segmentation results in Fig. \ref{fig:cmp_Monet}. Note that MONet dissects the black-and-white panda into different channels based on color, while our method successfully detects it without being biased by its color.

\begin{figure}[!t]
    \centering
    \includegraphics[width=0.7\linewidth,trim={0 0 0 0.1cm},clip]{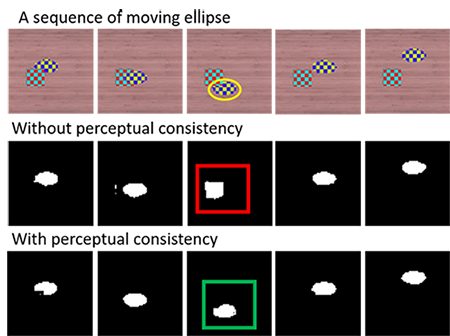}
    \caption{ {\it Effectiveness of the Perceptual Cycle-Consistency}: Note that an object will be assigned to different channels of the segmentation network from time to time (second row), showing temporal incoherence in the spatial disentanglement, however, the proposed perceptual cycle-consistency eliminates this incoherence, making the status of objects trackable.}
    \vspace{-0.3cm}
    \label{fig:Perceptual Consistency}
\end{figure}

{\bf Perceptual Cycle-Consistency}
We expect our model to exhibit the perceptual consistency mentioned in Section \ref{sec:method}, which means that in a temporally coherent sequence, each segmentation channel should keep track of the same object without identity switching. To verify the effectiveness of the perceptual cycle-consistency constraint, we train two segmentation networks on the Multi-Texture dataset, one with perceptual cycle-consistency but not the other. Then, we generate 256 sample sequences with varying positions and occlusions. Fig. \ref{fig:Perceptual Consistency} compares the two networks' behavior by visualizing their first output channels on the sequence in the top row. Without perceptual cycle-consistency, the first output channel mainly detects the ellipse but can switch to the square occasionally, especially when the two objects come close. However, with the perceptual cycle-consistency constraint enabled, the segmentation network can have each output channel focus on a fixed target throughout the whole sequence, with no identity switching. We evaluate the two network's performance by counting the number of target switches, shown in Table \ref{tb:quantitative_cyc}, which verifies that the proposed consistency constraint improves the temporal coherence of the generative factors.

\begin{figure}[!ht]
\begin{center}
  \includegraphics[width=0.48\textwidth]{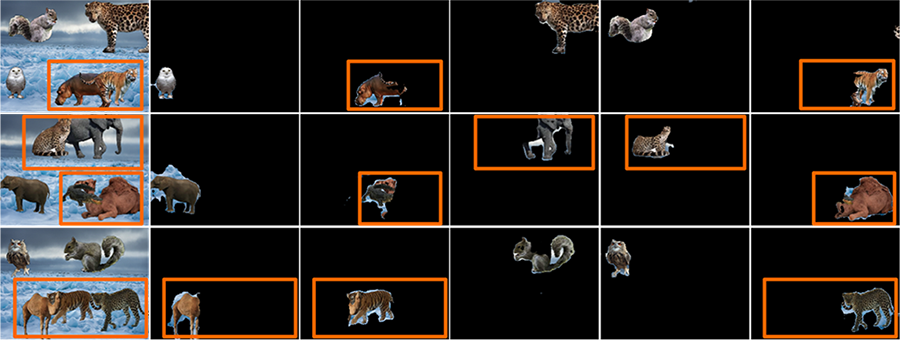}
\end{center}
\vspace{-0.3cm}
\caption{ {\it Occlusion affects the accuracy of the spatial disentanglement on the Flying Animals dataset.} Orange boxes highlight the regions where occlusion happens and the affected objects. }
\vspace{-0.3cm}
\label{fig:occlusion-flying-animal}
\end{figure}

\vspace{-0.05cm}
\section{Discussion}
\vspace{-0.1cm}

The evaluation of a method that aims at ``disentanglement'' is subjective, since we do not know what the model will be used for: It is common to hope that the hidden variables correspond to known components of the image-formation process, such as pose, scale, color, and shape. However, making that a quantitative benchmark may be misleading since, if that were the goal, we would simply capture those factors explicitly, for instance, via a conditional generative model. What we do observe is that the perceptual cycle-consistency, explicitly enforced in our model, enables the persistence of the representation, so identities of objects are not switched in different views. This would enable temporal consistency when the model is used as a prior in a sequential setting as shown in Fig. \ref{fig:Perceptual Consistency}.

Our model has limitations. The use of a VAE forces a hard trade-off between capacity, which affects the quality of the reconstructed image, and disentanglement, which is forced by the bottleneck. For complex scenes, there may not be a broad range of the trade-off parameter over which the model both captures the image statistics faithfully, and separates the hidden factors. Another limitation is the power of the inpainting model. For highly textured or complex scenes, the un-occluded region requires capturing the fine-grained context at a level of granularity higher than what our model affords, which may make it difficult for the segmentation network to learn perfect segmentation when occlusion happens, as shown in Fig. \ref{fig:occlusion-flying-animal}.

\section*{Acknowledgements}

Research supported by ONR N00014-17-1-2072 and N00014-19-1-2229.

{\small
\bibliographystyle{ieee_fullname}
\bibliography{egbib}
}

\end{document}